\documentclass{article}
\usepackage{booktabs} 
\usepackage{tikz}
\usepackage{amsmath}
\usepackage{pgfplots}
\usetikzlibrary{graphs} 
\usepackage{url}
\usepackage[textwidth=16.5cm]{geometry}

\title
      {Predicting Friction System Performance with Symbolic Regression and Genetic Programming with Factor Variables\footnote{Preprint of G. Kronberger, M. Kommenda, A. Promberger, F. Nickel. "Predicting friction system performance with symbolic regression and genetic programming with factor variables". In Proceedings of the Genetic and Evolutionary Computation Conference, pp. 1278-1285. ACM. (July 2018). The final publication is available at: https://dl.acm.org/doi/10.1145/3205455.3205522}}

 \author{Gabriel Kronberger, Michael Kommenda\\
   University of Applied Sciences Upper Austria\\School for Informatics, Communication, and Media\\
   Softwarepark 11, 4232 Hagenberg, Austria\\
   gabriel.kronberger@fh-ooe.at\\ \\
   Andreas Promberger, Falk Nickel\\
   Miba Frictec GmbH\\
   Peter-Mitterbauer-Str. 1, 4661 Roitham, Austria 
 }

\begin{document}
\maketitle
\begin{abstract}

    Friction systems are mechanical systems wherein friction is used
    for force transmission (e.g. mechanical braking systems or
    automatic gearboxes). For finding optimal and safe design
    parameters, engineers have to predict friction system
    performance. This is especially difficult in real-worlds applications, 
    because it is affected by many parameters. 

    We have used symbolic regression and genetic programming for
    finding accurate and trustworthy prediction models for this
    task. However, it is not straight-forward how nominal variables
    can be included. In particular, a one-hot-encoding is
    unsatisfactory because genetic programming tends to remove such
    indicator variables. We have therefore used so-called factor
    variables for representing nominal variables in symbolic
    regression models. Our results show that GP is able to produce symbolic regression models for predicting friction performance with predictive accuracy that is comparable to artificial neural networks. The symbolic regression models with factor variables are less complex than models using a one-hot encoding. 
\end{abstract}

\section{Introduction}

    Friction systems are mechanical systems wherein friction is used
    for force transmission between several moving components of the
    system (e.g. mechanical braking systems or automatic
    gearboxes). For the design of these systems, engineers have to find
    an optimal configuration which fulfills given specifications. In this
    process, they have to choose a friction material and
    lubricant type and calculate the necessary dimensions of the
    system components. Prediction models for friction system
    performance can facilitate this process, because many viable
    alternatives can be explored and compared based on their predicted
    performance, and necessary system dimensions can be
    calculated based on predicted friction performance. The
    alternative to using prediction models is to manufacture
    prototypes (physical models) of viable designs and performing
    tests with these prototypes. Overall, the effort to find a safe,
    durable and cost-efficient design can be reduced when using
    accurate prediction models for friction performance.

    Friction is a physical phenomenon that arises from forces acting
    between surfaces that are moving against each other. The overall
    friction force depends on the pressure on the two surfaces and the
    friction coefficient. The friction coefficient is mainly
    determined by the fine-grained structure of the surfaces and the
    type and physical characteristics of the lubricant. However,
    friction performance also depends on many other (dynamic) factors
    such as the sliding velocity, the pressure, the temperature of the
    surfaces and the lubricant and many more. Especially, in
    real-world applications there are many influence factors
    which makes it very difficult to build an accurate mathematical
    model derived from physical principles.  As a consequence,
    empirical modeling techniques have been used to find relevant
    correlations based solely on recorded data \cite{berger2002,hosenfeldt2014,
      Aleksendric201561,lughofer2016}. These methods can however only
    be used if enough high-quality data is available to detect all
    relevant correlations and for fitting the statistical model.

    \subsection{Previous Work}
Several different supervised learning techniques have been used for
empirical modeling of friction systems \cite{Ricciardi2017}. Besides
mathematical models derived from physical principles, artificial neural
networks (ANN) seem to be especially popular for this task. For example ANNs have
been used to predict wear of friction materials
\cite{Aleksendric2010,Aleksendric201561}, for optimizing braking performance as a component within a
neuro-genetic system 
\cite{cirovic2014}, for the prediction of the coefficient of friction
of different materials \cite{Senatore2011,Yang2013}, and for
predicting the tribological behavior of friction system
\cite{hosenfeldt2014}. Symbolic regression models and prediction models
based on fuzzy systems have been used to predict the coefficient of
friction, wear, temperatures, and noise-vibration-harshness ratings of
wet-application friction systems \cite{lughofer2016}. There,
the symbolic regression and fuzzy models produced higher
prediction errors than random forests, gradient boosted trees, or
support vector machines. However, in \cite{lughofer2016} the authors argue that
symbolic regression and fuzzy models still have merit, because both
methods produce models with much simpler structure and much fewer
parameters in comparison to random forests or gradient boosted trees.

\subsection{Motivation}
Our main motivation for the present research is that based on our
background knowledge of the physics of friction, we know that a given
friction system design always responds similarly to changes of load
parameters regardless of the specific type of material or
lubricant. For example, the temperature of the system will increase
when the amount of mechanical energy put into the system is
increased. Of course the actual temperature will be strongly dependent
on the specific material and lubricant that is used. However, the
overall correlations will be similar over all materials and
lubricants. This increase in temperature is just one example for a
known physical dependency, there are many more similar dependencies
that describe the overall behavior of friction systems.

A specific difficulty in this task is that we often do not know
details about the specific properties of materials or oils as
numeric features. Instead, we are only given nominal variables for the
material type or lubricant type. However, since these have a strong
effect on performance this information must be included in the model.

Correspondingly, a kind of a hierarchical empirical model should work
well for this particular application. A hierarchical model should
describe the general behavior of the overall system on a higher level
and the specifics for different materials or lubricants on a lower
level.  In previous empirical modeling efforts especially those using
artificial neural networks, this background knowledge about the
physical properties of friction systems has not been considered.  We
argue that with symbolic regression and genetic programming it is
rather easy to create such a hierarchical model using so-called \emph{factor
variables}. Factor variables allow to parameterize a global model
whereby the parameters depend on the observed values of nominal
variables.

Our aim in this article is to describe how factor variables can be
implemented for genetic programming and to demonstrate that this
approach is indeed capable of finding a general model structure and
parameter sets dependent on nominal variables. Finally, we analyze
whether it is possible to improve models for predicting friction
system performance.

\section{Methods}
In the following, we give details on the implementation of factor
variables for genetic programming. A simple problem is used to test
the implementation and demonstrate the feasibility of our
approach.

Next, we describe the procedure of data preparation for prediction
modeling. To check the correctness of our data preparation procedure
and to calculate baselines for the accuracy of our symbolic regression
models we train an ANN as well as a random forest using our
dataset. 
Finally, we test our proposed approach of symbolic regression with
factor variables as well as the straight-forward approach using a
one-hot encoding for nominal variables. The results of these two
approaches are discussed in detail to conclude this paper.

\subsection{Nominal Variables in Symbolic Regression}
   Nominal variables cannot usually be used directly in GP
   systems. Instead, it is necessary to encode nominal values to map
   them to numbers. Probably the easiest way is to use a so-called
   one-hot encoding, whereby binary indicator variables are included
   to represent each of the possible values of nominal variables.  A
   drawback of the one-hot encoding is that (1) it increases the
   number of variables in the dataset significantly, and (2) there is
   no guarantee that the indicator variables are actually included in
   the symbolic regression models. Instead, GP tends to remove
   indicator variables for nominal variable values.  
    
   For example, if we want to find a prediction model for the friction
   performance of twenty different materials, we have to add
   twenty additional binary variables, even if we originally only
   consider three numeric variables (pressure, sliding velocity, and
   temperature). Subsequently, it is necessary to increase the limit
   for the maximal size of models so that it is possible to include
   the indicator variables for materials. However, there is no
   guarantee that GP actually includes all indicator variables for
   materials. Instead, it will only include variables for materials
   which differ most from the average over all materials. For the
   remaining materials the prediction model will produce the same
   (average) value. Interpretation of the resulting symbolic
   expression will be difficult because references to material
   indicators will be scattered all over the model; in one part of the
   expression there might be references to materials A and B while in
   another part of the formula A and C are referenced (an example is
   given in Figure \ref{fig:model-one-hot2}). 

    What we aim for is a model where we can directly compare
    \emph{all} materials. It should be possible to compare how
    performance is changed over all materials as an effect of changing
    other parameters such as the pressure or temperature. Generally,
    GP should include a nominal variable if it has a strong impact on
    the output. If a nominal variable is included, the model should
    produce an output for all possible values of the nominal
    variable. In the resulting symbolic expression, references to
    nominal variables can be used in the same way as references to
    numeric variables. We call references to nominal variables
    \emph{factor variables}. For example it should be possible to find
    a linear model with material-specific factors as shown in Equation \ref{eqn:linear-model}.
    \begin{equation}
      f(x, Material) = \text{Material} \cdot  x + \text{Material}
      \label{eqn:linear-model}
    \end{equation}
    
    As stated above, we know that the
    overall behavior of the friction system is more or less the same regardless of
    the material or lubricant. For example the friction coefficient
    decreases when the temperature is increased regardless of the
    material. However, the absolute value or the rate of the decline
    of the friction coefficient certainly depends on the
    material. Thus, from the physical understanding of the system we
    know that the system follows a general behavior, reacting
    similarly to changes of input parameters for all materials,
    whereby the details might vary with different
    materials. Accordingly, we aim to find a global model structure
    that can be used to describe the whole system and at the same time
    identify the specific numeric parameters to fit the model
    parameters specifically to the observed system outputs for each
    material. The proposed factor variables represent exactly these
    material-specific numeric parameters and distinguish themselves from
    numeric model parameters the return the same value for all materials.

    This approach is conceptually very similar to non-linear
    regression using a predefined model structure and fitting the
    model independently to separate datasets containing only data for
    a specific material. The important difference is that GP with
    factor variables is able to consider the complete dataset while
    trying to find a globally valid model structure automatically.

\subsection{Implementation of Factor Variables for Genetic Programming}
    It is relatively easy to extend any GP system that is
    capable to solve symbolic regression problems to support factor
    variables. In the following, we assume a tree-based GP
    implementation which supports numeric variables and random
    constants as terminal symbols. The system must be capable of
    evolving or fitting numeric constants either via some form of
    mutation  or via specialized numerical
    optimization techniques \cite{topchy2001,kommenda2013effects}.
    \paragraph{Representation}
    A new type of symbol must be implemented and added to the
    terminal set. For a given dataset, the set of numeric variables
    and the set of nominal variables must be determined.  Additionally, 
    the set of possible values has to be determined and stored for each nominal variable. 

    \paragraph{Initialization of Random Trees}
    The routine for creation of random trees must be extended to
    randomly initialize the parameters for factor variables. When a
    terminal node for a factor variable is created, the routine must
    first select one variable from the set of nominal
    variables. Next a real-valued vector must be initialized, which
    maps each possible nominal to a numeric value (factor values). The
    reference to the nominal variable as well as the factor values are
    stored as data elements of the tree node.

    \paragraph{Mutation}
    If there is no other mechanism for fitting the numeric parameters
    of the model then the routine for mutating trees should be adapted
    to allow small changes to the factor values. We have found that
    adding a small normally distributed random value to a single
    factor value or to the whole vector works well.

\begin{figure}
\parbox{.45\linewidth}{
        \centering
\begin{tikzpicture}
  \node { + }
child {node { \begin{tabular} {lll} \multicolumn{2}{c}{$c$} \\ $A:$ & $1.0$ \\ $B:$ & $2.0$ \\ $C:$ & $1.5$  \end{tabular}  }}
child {node { * }
child {node { $x$ }}
child {node { \begin{tabular} {lll} \multicolumn{2}{c}{$c$} \\ $A:$ & $1.0$ \\ $B:$ & $2.0$ \\ $C:$ & $1.0$  \end{tabular}  }}
};
\end{tikzpicture}
}
\hfill 
\parbox{.45\linewidth}{
  \centering
  \begin{tabular}{l|l|c}    
  $x$ & $c$ & $f(x,c)$  \\
  \hline
  $3.0$ & $A$ & $4.0$ \\
  $2.0$ & $B$ & $6.0$ \\
  $1.0$ & $C$ & $2.5$ 
\end{tabular}
}

      \caption{\label{fig:factor-example}Example for an expression tree for a linear model with two factor variables referencing the same nominal variable $c$. The table shows the values of the expression for three different input vectors.}
    \end{figure}
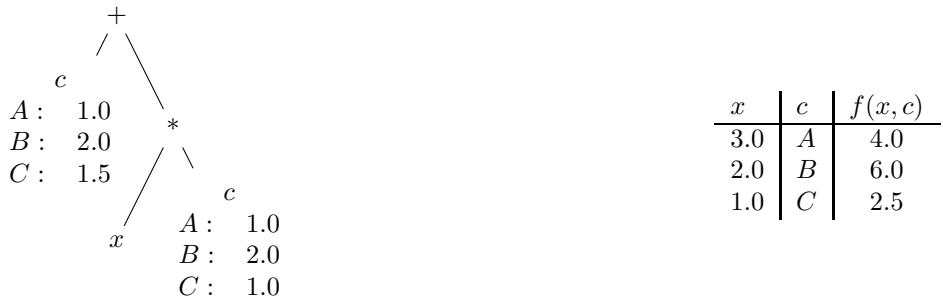

    \paragraph{Evaluation}
    Factor variables are evaluated to the value which corresponds to
    the observed value of the nominal variable.  An example for a tree
    with a factor variable as well as the value of the expression for
    three input vectors is shown in Figure \ref{fig:factor-example}.
    When evaluating a factor variables for a given row we look up the
    value of the referenced nominal variable and for this value the
    matching numeric value in the tree node is returned. For example,
    for the left node in the tree we would return the value $1.0$ if
    the value of $c$ is $A$ and $1.5$ if the value is $C$.
    
    \paragraph{Memetic optimization of numeric parameters}
    It is possible to rely solely on mutation to fit the numeric
    parameters to the available data. An alternative is to use guided
    techniques for fitting numeric constants, such as simple gradient
    descent or the L-BFGS algorithm \cite{liu1989limited}. When factor variables
    are included in the representation, the number of numeric
    parameters that need to be optimized is significantly
    increased. Thus, is it recommended to use a routine for optimizing
    constants before evaluating trees. These methods rely on gradient
    information, so it is necessary to be able to evaluate the partial
    derivatives of the loss function and therefore of the
    model itself for all numeric parameters and evaluate them for all
    observations in the dataset. Luckily, it is not necessary to
    calculate the derivative of the function symbolically. Instead,
    the partial derivatives can be evaluated together with the
    function output using automatic differentiation (cf. \cite{Rall1981ADT}). 
    
    Table \ref{tab:auto-diff} shows the results when calculating the
    gradient for the parameter vector $\theta$ and vectors $x$ and $c$. The elements of the
    vector $\theta$ are the same as in the tree in Figure
    \ref{fig:factor-example}. The $\theta_1\dots\theta_3$ stem from the
    left leaf of the tree in Figure \ref{fig:factor-example}, $\theta_4\dots\theta_6$ from the right leaf node.
    \begin{table}
      \centering
      \[ \theta = ( 1.0, 2.0, 1.5, 1.0, 2.0, 1.0 ) \]
      
  \begin{tabular}{l|l||c|c}    
    $x$ & $c$ & $f(x,c)$  & $\frac{\delta f}{\delta \theta}(x,c)$ \\
\hline
    $3.0$ & $A$ & $4.0$ & $(1, 0, 0, 3, 0, 0)$ \\
    $2.0$ & $B$ & $6.0$ & $(0, 0, 1, 0, 0, 2)$ \\
    $1.0$ & $C$ & $2.5$ & $(0, 1, 0, 0, 1, 0)$ \\
  \end{tabular}
  \caption{\label{tab:auto-diff}Calculation example for the partial
    derivatives $\frac{\delta f}{\delta \theta}(x,c)$ for the linear model shown in Figure \ref{fig:factor-example}. $\theta$ is the vector of the parameters values shown in the tree.
    Partial derivatives are necessary for optimizing $\theta$.}
    \end{table}

    Therefore, it is possible to optimize numeric parameters of
    symbolic regression models with factor variables using information
    about the gradient of the loss function which can be calculated
    efficiently in $O(nk)$ time using automatic differentiation where
    $n$ is the number of data points and $k$ is the number of
    parameters to be optimized.

    We use a squared error loss function and the Levenberg-Marquardt
    algorithm (LM) \cite{Levenberg1944} for optimization of numeric parameters in all
    experiments discussed in this paper. For each model that is
    evaluated by GP we first perform ten LM iterations. If the squared
    error can be reduced within those ten iterations the model in the
    population is updated with the optimized parameters. The number of
    iterations is set to such a small number because running LM to
    convergence for each model would incur a large runtime cost. It
    has been shown that a small number of iterations combined with
    writing back the parameters to the population is sufficient
    \cite{kommenda2013effects}.

 \subsection{A Synthetic Example for Factor Variables}
    To demonstrate the applicability of factor variables we have
    prepared a synthetic dataset using a function with a non-trivial
    shape which can be parameterized to take on different shapes shown
    in Equation \ref{eqn:synthetic-problem}. The parameterizations for four different
    values of the nominal variable $c$ are shown in the table below.
    \begin{figure}
      \centering
      \begin{equation}
        f(x) = \theta_{c,1} \exp \left(-0.08 \cdot x\right) - \exp \left( \theta_{c,2} \cdot x \right) - 0.1
        \label{eqn:synthetic-problem}
      \end{equation}
      
      \begin{tabular}{c|ccc}
        $c$ & $ \theta_1 $ & $ \theta_2 $ \\
        \hline 
        $A$ & $ 1.0 $ & $ -0.16 $  \\
        $B$ & $ 1.0 $ & $ -0.32 $  \\
        $C$ & $ 1.5 $ & $ -0.80 $  \\
        $D$ & $ 2.0 $ & $ -1.60 $  \\
      \end{tabular}
      \caption{\label{fig:synthetic-problem}A non-linear function with
        different parameterization depending on the variable $c$. We use this function to show the feasibility of genetic programming with factor variables.}
    \end{figure}

    To generate a dataset for training we sampled the function on a
    regular grid over $x$ for all four values of $c$. The sampled
    points are shown in Figure \ref{fig:model-output}. Using
    standard settings for tree-based genetic programming with the
    extensions for factor variables and gradient-based optimization of
    constants we found a perfect model in only a few generations.  The
    resulting expression is shown in Equation \ref{eqn:model-syn}. It
    can be seen that GP was able to find a correct model structure and
    parameters. The identified expression in Figure
    \ref{eqn:model-syn} can be simplified to a form that matches
    Equation \ref{eqn:synthetic-problem}.
    
    \begin{figure}
      \centering
        \begin{align}
\hat{f}(x) & =  c_{0}  + c_{1}  \cdot \exp \left( \text{x} \cdot c_{2}  \right)  + c_{3}  \cdot \exp \left( c_{4}  \cdot\text{x} \right)  \cdot \exp \left( c_{5}  \cdot\text{x} \right)  + c_{6}
\label{eqn:model-syn}
        \end{align}

\begin{tabular}{cc|cc}
  $c_{0}$ & $-0.25744$ & $c_{3, \text{c}=\text{A}}$ & $1.0$\\
  $  c_{1}$ & $-1.0$       & $c_{3, \text{c}=\text{B}}$ & $1.0$\\
  $c_{2, \text{c}=\text{A}}$ & $-0.16$ & $c_{3, \text{c}=\text{C}}$ & $1.5$\\
  $c_{2, \text{c}=\text{B}}$ & $-0.32$ &$c_{3, \text{c}=\text{D}}$ & $2.0$\\
  $c_{2, \text{c}=\text{C}}$ & $-0.8$   &$c_{4}$ & $-0.034663$\\
  $c_{2, \text{c}=\text{D}}$ & $-1.6$   & $c_{5}$ & $-0.045337$\\
  & & $c_{6}$ & $0.15744$
\end{tabular}

\caption{Symbolic regression model with factor variables found by GP for the training dataset sampled from the synthetic problem shown in Figure \ref{fig:synthetic-problem}. The identified expression can be rearranged to match the original function (Equation \ref{eqn:synthetic-problem}) exactly.}
        \end{figure}

    Figure \ref{fig:model-output} shows the outputs of the
    model shown in Equation \ref{eqn:model-syn} as
    well as the data points that have been used for training. The
    model interpolates perfectly between the sampled training points.
    
    \begin{figure}
      \centering
      \begin{tikzpicture}
       \begin{axis}[
xlabel=$x$,
ylabel=$f(x)$,
         ]
         \legend{train (c=A), train (c=B), train (c=C), train (c=d), pred (c=A), pred (c=B), pred (c=C), pred (c=D)}
         \addplot+[only marks,color=red,mark=x, each nth point={2}, filter discard warning=false
         ]  table[x=x,y=y] {models/solution-osgpfc_syn-1_output_a.txt};
         \addplot+[only marks,color=red,mark=+, each nth point={2},filter discard warning=false
         ]  table[x=x,y=y] {models/solution-osgpfc_syn-1_output_b.txt};
         \addplot+[only marks,color=red,mark=o, each nth point={2},filter discard warning=false
         ]  table[x=x,y=y] {models/solution-osgpfc_syn-1_output_c.txt};
         \addplot+[only marks,color=red,mark=*, each nth point={2},filter discard warning=false
         ]  table[x=x,y=y] {models/solution-osgpfc_syn-1_output_d.txt};
         \addplot+[color=red,mark=none,
         ]  table[x=x,y=model] {models/solution-osgpfc_syn-1_output_a.txt};
         \addplot+[color=red,mark=none,
         ]  table[x=x,y=model] {models/solution-osgpfc_syn-1_output_b.txt};
         \addplot+[color=red,mark=none,
         ]  table[x=x,y=model] {models/solution-osgpfc_syn-1_output_c.txt};
         \addplot+[color=red,mark=none,
         ]  table[x=x,y=model] {models/solution-osgpfc_syn-1_output_d.txt};
       \end{axis}
      \end{tikzpicture}
      \caption{\label{fig:model-output}Data points used for training as well as the predictions from the symbolic regression model (Equation \ref{eqn:model-syn}) for the synthetic problem . For this simple problem GP with factor variables produces perfect predictions.}
    \end{figure}
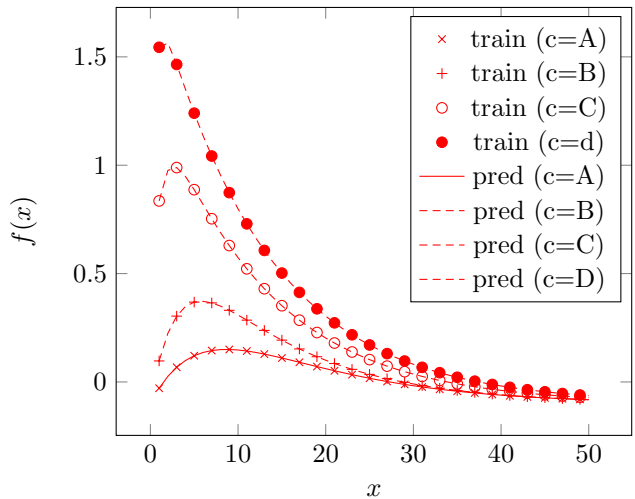

    This result demonstrates the merit of factor variables. In this
    example this is clearly beneficial as we know that there is a
    common model structure describing a family of models and only the
    parameterization depends on the values of nominal variables. To
    stress the same point again; the procedure is equivalent to
    non-linear least squares fitting with a given model structure and
    for four independent datasets for $A, B, C$, and $D$. However,
    since we assume that the correct model structure is unknown and
    should be identified it is worthwhile to combine all data.

    \section{Experiments}
    In the following we describe our experiments, in which we used GP
    with factor variables to produce prediction models for the performance of friction
    systems.
    
\subsection{Data Preparation}
    We use data published together with 
    \cite{lughofer2016}\footnote{available from \url{https://dev.heuristiclab.com/AdditionalMaterial}}.  This is a collection of data acquired
    over multiple years from standardized friction tests in an
    industrial setting. It contains data from tests for a diverse set of
    of materials, oils, and types of surface grooving.

    We filtered the data and kept only data from tests using the most
    frequently used oil type and grooving type. From the remaining
    rows we only kept data for the most commonly used materials. This
    filtering of the data is necessary because the original dataset
    does not contain data for all possible combinations of materials
    and oils even for a single grooving type. Therefore, it is not
    feasible to fit a model which contains interaction effects between
    materials and oils.  The filtered dataset contains data from tests
    for four different materials identified by the nominal values $A,
    B, C, D$.  In the testing procedure, a rotating disk is pressed
    against a fixed disk until the rotating disk is stopped (brake
    test). The starting velocity of the rotating disk is the same for
    all cycles in the test; the pressure is varied over the test. For
    each pressure level multiple cycles (repetitions) are
    performed. For each cycle, the average coefficient of friction is
    determined by measuring the time until the rotating disk is
    completely stopped.

    Consequently, the data allows to build a model to predict the
    coefficient of friction as a function of two numerical variables
    (pressure and the number of cycles performed) and one nominal
    variable (material). As we will see below, the number of cycles is
    relevant, because the material deteriorates over the test run
    because of wear, with the effect of a decreased friction
    performance.

    The numeric variables were both scaled to the range $[0..1]$.

    \subsection{Modeling}
    For prediction modeling we used four different techniques:
    artificial neural network, random forest, symbolic regression and
    symbolic regression with factor variables. Only the later
    can handle nominal variables directly. For the other algorithms we extended our
    dataset to include binary indicator variables for each of the four
    materials.  In the following we describe the algorithm
    configurations for our experiments in detail. The results are
    discussed in the next section.

    \paragraph{Artificial Neural Network}
    First, we trained an artificial neural network using our filtered
    dataset for a comparison of the accuracy of the models with
    previously published results. For the artificial neural network we
    used the implementation available in
    ALGLIB\footnote{\url{http://www.alglib.net}}. We used six input
    neurons (pressure, cycles, Mat=a, Mat=b, Mat=c, Mat=d) and only
    one hidden layer of ten nodes. This particular implementation
    allows to set a decay parameter for regularization of network
    weights. We used 10-fold cross-validation to determine the best
    value for the decay parameter ($\in \{1\times10^{-5},5\times10^{-5},1\times10^{-4},5\times10^{-4},1\times10^{-3},4\times10^{-3},1\times10^{-2},5\times10^{-2}, 0.1\}$).

    \paragraph{Random Forest}
    As reported in \cite{lughofer2016}, the random forest model performed very well; often producing the best predictions. We therefore also
    trained a random forest model with the filtered dataset using ALGLIB
    and the same set of features as for ANN. The number of trees was set to 200. The parameter $M$ was set
    to $0.5$; therefore, for each tree half of the features were
    selected randomly). The parameter $R$ that determines how many
    rows are selected randomly was determined through 10-fold
    cross-validation ($R\in \{0.1, 0.2, 0.3, 0.4, 0.5, 0.6\}$).

    \paragraph{Symbolic Regression}
    For symbolic regression we used genetic programming with and
    without factor variables\footnote{We used the GP implementation in HeuristicLab \url{http://dev.heuristiclab.com}}. In both cases, we used a population size
    of 200 trees initialized randomly using the PTC2 algorithm
    \cite{Luke2000}. Number of generations was fixed to 100. For the
    function set we used $\{+, -, *, /, \log(), \exp() \}$. The
    fitness function is the mean of squared errors for both configurations. We do not apply grouping by factor values for the calculation of the fitness criterion. A grouping by factors is not necessary because with factor variables the symbolic regression solutions can be evaluated on the complete dataset. 
    In both configurations
    the numeric parameters of all models are optimized before fitness
    evaluation using ten iterations of LM. 75\% of the data were used
    for training. The best model on the training set is evaluated on
    the remaining data. For the runs with factor variables we set the
    limit for the size of trees to 25 nodes because we aim to find
    rather simple models. For the experiments with one-hot encoding of
    materials we increased the limit to 50 nodes to allow inclusion of
    all indicator variables. When nominal variables can be referenced
    directly we have three different variables, with the one-hot
    encoding the dataset is increased to six
    variables. Correspondingly, we also doubled the limit for the tree
    size.
    
    \section{Results}
    Table \ref{tab:results} shows the prediction errors for all tested
    methods.     
    \begin{table}
      \centering
      \begin{tabular}{lc}
        Model & average relative error \\
        \hline
        Linear regression & 4.15 \% \\
        Artificial neural network & 2.86 \%  \\ 
        Random forest & 3.08 \% \\ 
        Sym. reg. with one-hot-encoding & 2.77 \% \\ 
        Sym. reg. with factor variables & 2.84 \% \\ 
    \end{tabular}
    \caption{\label{tab:results}Overview of the modeling results. Linear regression is not able to capture the non-linear effect of the pressure and therefore produces worse predictions compared to the other techniques.}
    \end{table}
    
    Overall, the results for all methods are similar. All models
    have an average prediction error which is well below 5\%. The difference between the linear model with an error of 4\% and the best models (SR and ANN) with errors around 2.7\% is huge from a practical point of view. The larger prediction error produced by the linear model is a consequence of the fact, that the linear model cannot capture the assumed non-linear effect of the pressure parameter. This is in line with previous results \cite{lughofer2016, hosenfeldt2014} and the expected behavior of the studied friction systems. This assumed non-linearity was the initial motivation for us to apply GP to this particular problem.

    Our
    results for this dataset are slightly better than reported in
    \cite{lughofer2016} where prediction errors for the
    coefficient of friction between 5\% and 6\% are reported. Our results are also
    significantly better than the results reported by
    \cite{hosenfeldt2014}, where the authors report an deviation of 8\% and
    state that the measurement error with reference to friction is
    around 5\%.

\begin{figure*}
      \centering
      \begin{tikzpicture}
       \begin{axis}[
xlabel={pressure},ymin=0.05,ymax=0.13,
ylabel=$\mu_{\text{avg.}}$,
width=5cm,
cycle list name=linestyles,
         title={Mat=a}]
         \addplot+[smooth,mark=none,color=black,
         ]  table[x=p_spec,y=model] {models/solution-nn-cf_over_p_a.txt};
         \addplot+[const plot mark mid,mark=none,color=black,
         ]  table[x=p_spec,y=model] {models/solution-rf-cf_over_p_a.txt};
         \addplot[only marks,mark=x,color=black,
         ]  table[x=x2,y=cf] {data/cf_brake_a.txt};
       \end{axis}
      \end{tikzpicture}
      \begin{tikzpicture}
       \begin{axis}[
           xlabel={pressure},ymin=0.05,ymax=0.13,yticklabels={},
           width=5cm,
         cycle list name=linestyles,title={Mat=b}]
         \addplot+[smooth,mark=none,color=black,
         ]  table[x=p_spec,y=model] {models/solution-nn-cf_over_p_b.txt};
         \addplot+[const plot mark mid,mark=none,color=black,
         ]  table[x=p_spec,y=model] {models/solution-rf-cf_over_p_b.txt};
         \addplot[only marks,mark=x,color=black,
         ]  table[x=x2,y=cf] {data/cf_brake_b.txt};
       \end{axis}
      \end{tikzpicture}
      \begin{tikzpicture}
       \begin{axis}[
xlabel={pressure},ymin=0.05,ymax=0.13,yticklabels={},
width=5cm,
         cycle list name=linestyles,title={Mat=c}]
         \addplot+[smooth,mark=none,color=black,
         ]  table[x=p_spec,y=model] {models/solution-nn-cf_over_p_c.txt};
         \addplot+[const plot mark mid,mark=none,color=black,
         ]  table[x=p_spec,y=model] {models/solution-rf-cf_over_p_c.txt};
         \addplot[only marks,mark=x,color=black,
         ]  table[x=x2,y=cf] {data/cf_brake_c.txt};
       \end{axis}
      \end{tikzpicture}
      \begin{tikzpicture}
       \begin{axis}[
xlabel={pressure},ymin=0.05,ymax=0.13,yticklabels={},
width=5cm, cycle list name=linestyles,title={Mat=d}]
         \addplot+[smooth,mark=none,color=black,
         ]  table[x=p_spec,y=model] {models/solution-nn-cf_over_p_d.txt};
         \addlegendentry{NN}
         \addplot+[const plot mark mid,mark=none,color=black,
         ]  table[x=p_spec,y=model] {models/solution-rf-cf_over_p_d.txt};
         \addlegendentry{RF}
         \addplot[only marks,mark=x,color=black,
         ]  table[x=x2,y=cf] {data/cf_brake_d.txt};
         \addlegendentry{Data}
       \end{axis}
      \end{tikzpicture}
      \begin{tikzpicture}
       \begin{axis}[
xlabel={cycles},ymin=0.05,ymax=0.13,
ylabel=$\mu_{\text{avg.}}$,
width=5cm,
cycle list name=linestyles]
         \addplot+[smooth,mark=none,color=black,
         ]  table[x=cyc,y=model] {models/solution-nn-cf_over_cyc_a.txt};
         \addplot+[const plot mark mid,mark=none,color=black,
         ]  table[x=cyc,y=model] {models/solution-rf-cf_over_cyc_a.txt};
         \addplot[only marks,mark=x,color=black,
         ]  table[x=cycles,y=cf] {data/cf_brake_a.txt};
       \end{axis}
      \end{tikzpicture}
      \begin{tikzpicture}
       \begin{axis}[
           xlabel={cycles},ymin=0.05,ymax=0.13,yticklabels={},
           width=5cm,
         cycle list name=linestyles]
         \addplot+[smooth,mark=none,color=black,
         ]  table[x=cyc,y=model] {models/solution-nn-cf_over_cyc_b.txt};
         \addplot+[const plot mark mid,mark=none,color=black,
         ]  table[x=cyc,y=model] {models/solution-rf-cf_over_cyc_b.txt};
         \addplot[only marks,mark=x,color=black,
         ]  table[x=cycles,y=cf] {data/cf_brake_b.txt};
       \end{axis}
      \end{tikzpicture}
      \begin{tikzpicture}
       \begin{axis}[
xlabel={cycles},ymin=0.05,ymax=0.13,yticklabels={},
width=5cm,
         cycle list name=linestyles]
         \addplot+[smooth,mark=none,color=black,
         ]  table[x=cyc,y=model] {models/solution-nn-cf_over_cyc_c.txt};
         \addplot+[const plot mark mid,mark=none,color=black,
         ]  table[x=cyc,y=model] {models/solution-rf-cf_over_cyc_c.txt};
         \addplot[only marks,mark=x,color=black,
         ]  table[x=cycles,y=cf] {data/cf_brake_c.txt};
       \end{axis}
      \end{tikzpicture}
      \begin{tikzpicture}
       \begin{axis}[
xlabel={cycles},ymin=0.05,ymax=0.13,yticklabels={},
width=5cm, cycle list name=linestyles]
         \addplot+[smooth,mark=none,color=black,
         ]  table[x=cyc,y=model] {models/solution-nn-cf_over_cyc_d.txt};
         \addlegendentry{NN}
         \addplot+[const plot mark mid,mark=none,color=black,
         ]  table[x=cyc,y=model] {models/solution-rf-cf_over_cyc_d.txt};
         \addlegendentry{RF}
         \addplot[only marks,mark=x,color=black,
         ]  table[x=cycles,y=cf] {data/cf_brake_d.txt};
         \addlegendentry{Data}
       \end{axis}
      \end{tikzpicture}

      \caption{\label{fig:rf-and-nn-sensitivity}Training data and partial dependence plots for ANN and the RF models over pressure (first row) and the number of cycles (second row). The RF model produces a step-function typical for decision tree models. The ANN model produces a smooth interpolation. The plots show no signs of overfitting. Materials $c$ and $d$ have very similar performance. The data contains a set of measurements at the beginning and the end of the test procedure.}

    \end{figure*}
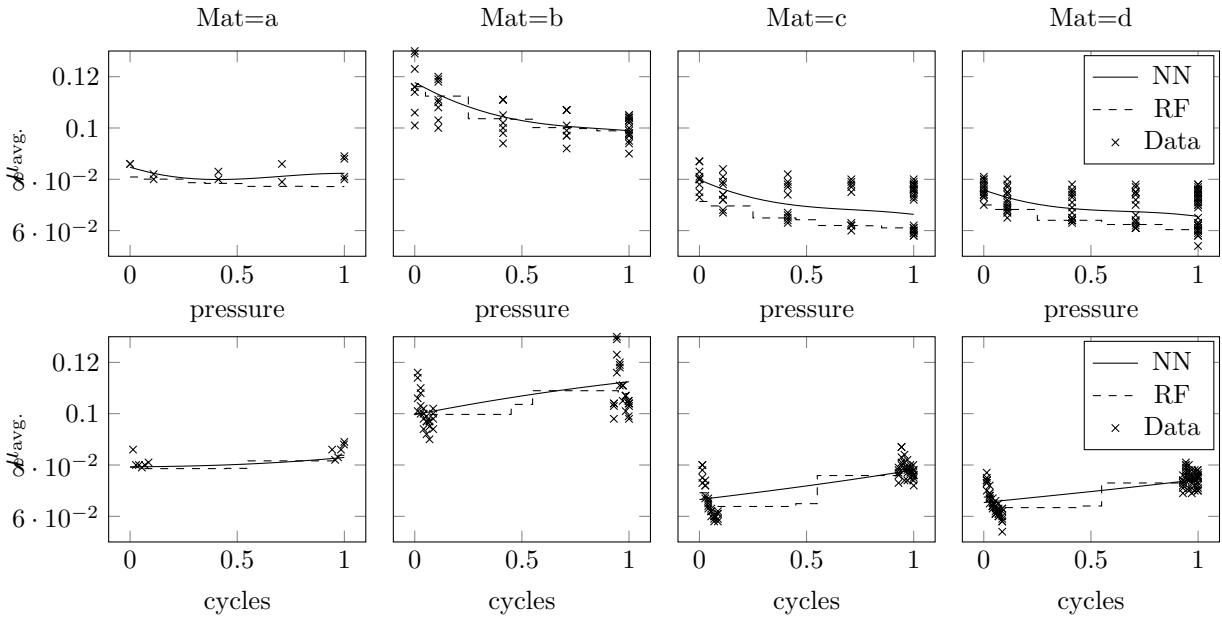

    As a consequence, we checked the plausibility of our
    results using partial dependence plots of the model outputs over
    the input variables.  Figure \ref{fig:rf-and-nn-sensitivity} shows
    the outputs of the ANN and the RF models for independent variation
    of pressure and the number of cycles. The training points are also
    shown for reference. In the plots in Figure \ref{fig:rf-and-nn-sensitivity}, the step-wise
    interpolation of the RF model becomes apparent. These discrete
    jumps in the predicted performance are highly
    problematic from a practical point of view because the underlying friction system does not have such jumpy behavior. Therefore, we cannot recommend the RF model for this
    application.  In contrast, the ANN produces a nice smooth
    interpolation. This might be a reason why ANNs are rather popular
    in empirical modeling of friction. The predictions produced by the
    ANN model for our data are smooth and plausible. The model behaves
    similar over the whole space of valid values for the pressure and
    the cycles. The charts do not give an indication of overfitting.

Figure \ref{fig:model-one-hot2} shows the resulting symbolic
regression model when a one-hot encoding for the material is used. It
is difficult to read and interpret the model because the references to
the binary indicator variables are scattered over the whole model. The
binary indicator variables have the effect that numeric parameters are
turned on or off depending on the observer material. For example, $c_2$
is only active for material $b$. Notably, GP has removed the variable
Mat=a because it found a model that produces the correct predictions
for material $a$ when all other material indicators equal to zero. The
output of the model is shown in Figure \ref{fig:symreg-sensitivity}. 

    \begin{figure}
  \begin{multline*}
\nonumber
\widehat{\mu_{avg}} = 
\cfrac{   c_{0}  \cdot\text{cycles} + c_{1}  \cdot p + c_{2}  \cdot\text{M=b} + \exp \left( c_{3}  \cdot\text{M=c} \right)  \cdot c_{4} }{\exp \left( c_{5}  \cdot\text{M=d} + c_{6}  \cdot\text{cycles} + c_{7}  \cdot p  \right)  \cdot c_{8} }  \\[10pt]
 + \exp \left( c_{9}  \cdot\text{M=d} + c_{10}  \cdot p \right)  \cdot c_{11} 
 + \cfrac{ c_{12}  \cdot\text{cycles} + c_{13}  \cdot\text{M=b} + c_{14}  \cdot p + c_{15}  \cdot\text{M=c} + c_{16}  \cdot\text{M=d}  }{  \exp \left( c_{17}  \cdot\text{M=b} + c_{18}  \cdot p \right)  \cdot  \left( c_{19}  \cdot p + c_{20}  \cdot\text{cycles} \right)  + c_{21} } + c_{22}
  \end{multline*}
\caption{\label{fig:model-one-hot2}Symbolic regression model for the coefficient of friction using one-hot encoding. \text{M=*} variables represent binary indicator variables for the four materials. Notably, the variable \text{M=a} does not occur in the model, meaning that the model produces the predictions for material $a$ if all material indicator variables have the value zero. The model is difficult to read because references to materials are scattered over the whole model. The coefficient values $c_1-c_{22}$ are not shown because of limited space.}
 \end{figure}

     Figure \ref{fig:model-factor} shows the symbolic regression model
     when using factor variables. In comparison to the model with
     one-hot encoding, this model is easier to read because the
     material-specific parameters are collected in three factor
     variables. However, the number of parameters is 22 for both
     models.  The output of the model with factor variables is also
     shown in Figure \ref{fig:symreg-sensitivity}. In comparison, all
     three models ANN, symbolic regression with one-hot encoding, and
     symbolic regression with factor variables are very similar. The
     two symbolic regression models are almost indistinguishable.
     
    \begin{figure}
\begin{multline*}
  \nonumber
  \widehat{\mu_{avg}}  = 
  \bigl( c_{0}  \cdot\text{cycles} + c_{1,\text{M}}  \cdot \exp \left( c_{2}  \cdot\text{cycles} +  c_{3}  \cdot p \right)\\
  + c_{4,\text{M}}  \cdot  \left( c_{5}  \cdot p + \exp \left( c_{6}  \cdot p \right)  + c_{7} \right)  \bigr) \cdot \frac{1}{c_{8}  \cdot p + c_{9,\text{M}}  + c_{10}}  + c_{11} 
\end{multline*}
      \caption{\label{fig:model-factor}Symbolic regression model with factor variables for the coefficient of friction. Factor variables are given as $c_{\cdot,M}$. This model is more readable than the model with one-hot encoding (\ref{fig:model-one-hot2}) because the material-dependent parameters are collected in three factor variables. The overall properties of the prediction function are thus easier to identify.}
    \end{figure}

    \begin{figure*}
      \centering
      \begin{tikzpicture}
        \begin{axis}[
            ymax=0.12, ymin=0.06,
xlabel=pressure,
ylabel=$\mu_{\text{avg.}}$,
cycle list name=linestyles
         ]
         \addplot+[smooth,mark=none,
         ]  table[x=p_spec,y=model] {models/solution-symreg-factors-cf_over_p_a.txt};
         \addlegendentry{SR factors}
         \addplot+[smooth,mark=none,
         ]  table[x=p_spec,y=model] {models/solution-symreg-one-hot_2-cf_over_p_a.txt};
         \addlegendentry{SR one-hot}
         \addplot+[smooth,mark=none,
         ]  table[x=p_spec,y=model] {models/solution-nn-cf_over_p_a.txt};
         \addlegendentry{ANN}

         \addplot+[smooth,mark=none,
         ]  table[x=p_spec,y=model] {models/solution-symreg-factors-cf_over_p_b.txt};
         \addplot+[smooth,mark=none,
         ]  table[x=p_spec,y=model] {models/solution-symreg-one-hot_2-cf_over_p_b.txt};
         \addplot+[smooth,mark=none,
         ]  table[x=p_spec,y=model] {models/solution-nn-cf_over_p_b.txt};

         \addplot+[smooth,mark=none,
         ]  table[x=p_spec,y=model] {models/solution-symreg-factors-cf_over_p_c.txt};
         \addplot+[smooth,mark=none,
         ]  table[x=p_spec,y=model] {models/solution-symreg-one-hot_2-cf_over_p_c.txt};
         \addplot+[smooth,mark=none,
         ]  table[x=p_spec,y=model] {models/solution-nn-cf_over_p_c.txt};

         \addplot+[smooth,mark=none,
         ]  table[x=p_spec,y=model] {models/solution-symreg-factors-cf_over_p_d.txt};
         \addplot+[smooth,mark=none,
         ]  table[x=p_spec,y=model] {models/solution-symreg-one-hot_2-cf_over_p_d.txt};
         \addplot+[smooth,mark=none,
         ]  table[x=p_spec,y=model] {models/solution-nn-cf_over_p_d.txt};

       \end{axis}
       \end{tikzpicture}
      \begin{tikzpicture}
       \begin{axis}[
            ymax=0.12, ymin=0.06,
xlabel=cycles,yticklabels={},
cycle list name=linestyles
         ]
         \addplot+[smooth,mark=none,
         ]  table[x=cyc,y=model] {models/solution-symreg-factors-cf_over_cyc_a.txt};
         \addplot+[smooth,mark=none,
         ]  table[x=cyc,y=model] {models/solution-symreg-one-hot_2-cf_over_cyc_a.txt};
         \addplot+[smooth,mark=none,
         ]  table[x=cyc,y=model] {models/solution-nn-cf_over_cyc_a.txt};

         \addplot+[smooth,mark=none,
         ]  table[x=cyc,y=model] {models/solution-symreg-factors-cf_over_cyc_b.txt};
         \addplot+[smooth,mark=none,
         ]  table[x=cyc,y=model] {models/solution-symreg-one-hot_2-cf_over_cyc_b.txt};
         \addplot+[smooth,mark=none,
         ]  table[x=cyc,y=model] {models/solution-nn-cf_over_cyc_b.txt};

         \addplot+[smooth,mark=none,
         ]  table[x=cyc,y=model] {models/solution-symreg-factors-cf_over_cyc_c.txt};
         \addplot+[smooth,mark=none,
         ]  table[x=cyc,y=model] {models/solution-symreg-one-hot_2-cf_over_cyc_c.txt};
         \addplot+[smooth,mark=none,
         ]  table[x=cyc,y=model] {models/solution-nn-cf_over_cyc_c.txt};

         \addplot+[smooth,mark=none,
         ]  table[x=cyc,y=model] {models/solution-symreg-factors-cf_over_cyc_d.txt};
         \addplot+[smooth,mark=none,
         ]  table[x=cyc,y=model] {models/solution-symreg-one-hot_2-cf_over_cyc_d.txt};
         \addplot+[smooth,mark=none,
         ]  table[x=cyc,y=model] {models/solution-nn-cf_over_cyc_d.txt};

       \end{axis}
      \end{tikzpicture}
      \caption{\label{fig:symreg-sensitivity}Partial dependence plots of the two symbolic regression models (Figures \ref{fig:model-one-hot2} and \ref{fig:model-factor}) and the trained ANN model. The two SR models are almost indistinguishable. The ANN model smooths slightly more strongly.}
    \end{figure*}
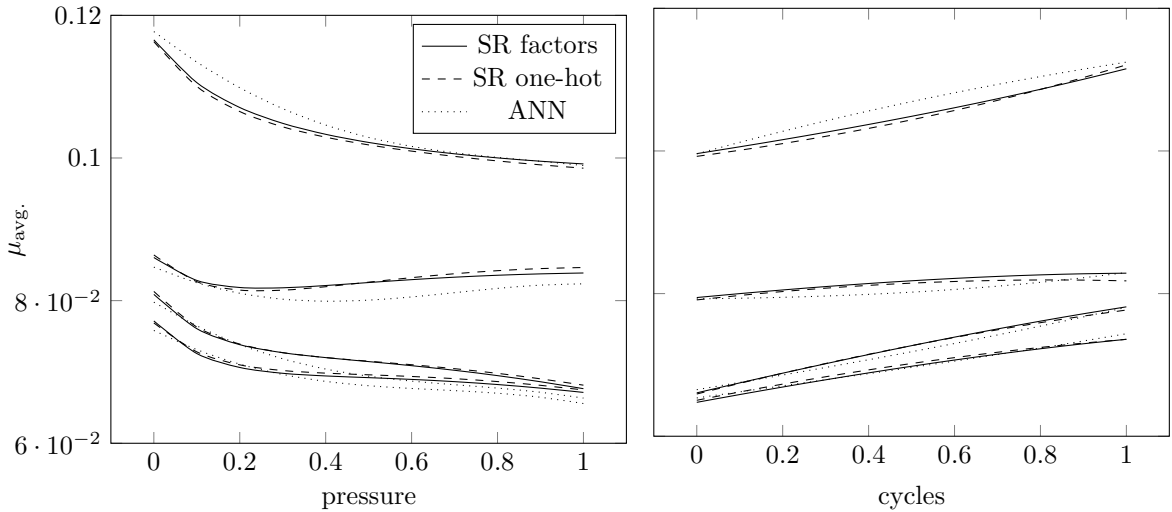

    Figure \ref{fig:symreg-sensitivity} shows the partial dependence plots of the trained ANN model, the symbolic regression models with one-hot encoding (Figure \ref{fig:model-one-hot2}), and the symbolic regression model with factor variables (Figure \ref{fig:model-factor}). The outputs and the overall accuracy of all models are very similar (cf. Table \ref{tab:results}). All three models clearly show the expected non-linear correlation with the pressure parameter. However, the symbolic regression model with factor variables is much easier to read and verify than the model using a one-hot encoding. The ANN model has two orders of magnitude more parameters and is a black-box model. 
    
    \section{Summary and Conclusions}
  We proposed an extension of genetic programming for symbolic
  regression which makes it possible to use nominal variables directly
  instead of using a one-hot encoding. 
    
  The results of our experiments show that it is possible to find
  accurate and trustworthy models for predicting friction system
  performance using genetic programming with factor variables. The
  symbolic regression models are more compact (number of nodes) than
  with one-hot-encoding, and easier to read because references to
  nominal variable value are collected within factor variables. The
  total number of parameters could however not be reduced and might be
  even increased because each factor variable contains one parameter
  for each possible value of the nominal variable.

  The usage of factor variables enforces that parameters for each
  nominal value are identified, in contrast to the one-hot encoding
  where some indicator variables might be removed by GP. As an
  additional benefit, the identified factor values can be compared
  directly to find for which nominal values the predictions are
  similar or dissimilar. This could be interesting to estimate the
  similarity of materials using the similarity in the parameter space
  for the symbolic regression model.

  Both symbolic regression models as well as the ANN produce very
  similar predictions for our friction dataset with average errors of
  around 2.8\%. From a practical point of view the difference between
  the linear model with an average error of 4\% and the best models
  with average errors of 2.8\% is huge.

  The comparison with techniques used in previous work for empirical
  modeling and prediction of friction system performance showed that
  the ANN model also produce very good predictions and smooth
  interpolations. However, ANN models are black-boxes and cannot be as easily inspected or verified as the SR models.
  The RF model is not satisfactory for this application because of its step-wise prediction function. Even tough the cross-validation error of around 3\% of the RF model is good, it cannot be
  recommended for the predicting friction performance in the design of friction systems.

  We have not yet studied how the usage of factor variables might
  increase the likelihood of overfitting. Especially, for datasets
  with nominal variables with many different values (e.g. more than
  10), overfitting might become an issue because of the large number of
  parameters that are introduced. Similarly, if a dataset contains multiple nominal variables, the data becomes increasingly sparse relative to the number of possible combinations of nominal, leading to problems when fitting a model. In this paper, we have used datasets with only a single nominal variable. Further research  is necessary
  to study the sensitivity of the method to a growing number of nominal variables in detail.

  In our experiments, we have used a gradient-based technique for the
  optimization of factor values. We have not yet analyzed whether this
  is strictly necessary, or whether the factor values can also be
  evolved directly e.g. through mutation. We strongly believe that it
  will be hard to find good factor values without gradient-based
  optimization, but we do not have enough evidence to support
  this. There might be a trade-off where a certain number of
  iterations is necessary to find reasonable values but too many
  iterations lead to overfitting because of the large number of
  parameters.
  
  \subsection*{Acknowledgements}
  G.K. gratefully acknowledges support by the Christian Doppler
  Research Association and the Federal Ministry for Digital and
  Economic Affairs within the \emph{Josef Ressel Centre for Symbolic
    Regression}.  M.K. gratefully acknowledges support by the Austrian
  Research Promotion Agency (FFG), the Federal Ministry for Digital and
  Economic Affairs, and the State Government of Upper Austria within
  the COMET Project \emph{Heuristic Optimization in Production and
    Logistics (HOPL), \#843532}.

  GK wrote the code and prepared and
  analyzed the experiments and wrote the paper. MK reviewed the code
  and analyzed the experiments. GK and MK came up with the idea of
  factor variables for symbolic regression together. AP and FN
  collected and prepared all the data and validated the models.

  \bibliographystyle{apalike}
  \bibliography{symreg-factors.bib}

\end{document}